\documentclass{article}
\usepackage[preprint]{neurips_2025}
\usepackage[utf8]{inputenc}
\usepackage[T1]{fontenc}
\usepackage{hyperref}
\usepackage{url}
\usepackage{booktabs}
\usepackage{amsfonts}
\usepackage{nicefrac}
\usepackage{microtype}
\usepackage{xcolor}
\usepackage{graphicx}
\usepackage{natbib}
\title{TinyTim: A Family of Language Models for Divergent Generation}
\author{%
  Christopher J. Agostino \\
  NPC Worldwide \\
  Bloomington, IN 47404 \\
  \texttt{info@npcworldwi.de}
}
\begin{document}
\maketitle
\begin{abstract}
In the search for artificial general intelligence, model development and training has focused primarily on vast datasets of known problems and their accepted solutions. This process necessarily produces convergent systems which are fundamentally incapable of the conceptual reframing that is required for genuine creative breakthroughs. Inspired by the divergent cognitive processes that allow humans to make such creative leaps, our work introduces a family of language models, TinyTim, to serve as sources of divergent generation within broader systems. These models have been created by fine-tuning on the anti-parsimonious text of James Joyce's `Finnegans Wake'. Quantitative analysis of both an unsupervised fine-tuned model (TinyTim-V1) and a new instruction-tuned variant (TinyTim-V2) demonstrates a profound capacity for lexical invention; the foundational V1 model exhibits a Yule's K score for lexical richness over twenty times greater than that of convergent baselines. This trait is a stable property of the family, as the instruction-tuned V2 maintains a statistically distinct profile and resists factual convergence, sacrificing benchmark performance to preserve its core generative style. This work establishes a methodology for engineering specialized divergent models that, when paired with convergent systems, can reframe problems and force breakthroughs beyond the reach of statistical optimization alone.

\end{abstract}
\section{Introduction}
Large Language Models (LLMs) based on the Transformer architecture \citep{vaswani2017attention} have demonstrated powerful capabilities in synthesizing statistically likely patterns from vast data \citep{brown2020language}. However, this strength is also a fundamental limitation, constraining them to convergent, `mean-reverting' outputs that inhibit the generation of genuinely novel hypotheses \citep{dweck2023creativity}. This issue reflects long-standing critiques of artificial reason, where formal systems struggle with the ambiguity and creative extrapolation inherent in human cognition \citep{dreyfus1992, piantadosi2022meaning}. 
Human creativity is often described by dual-process theories that posit an interplay between focused, convergent thinking for evaluation and associative, divergent thinking for ideation \citep{guilford1967nature, sternberg1985implicit}. This cognitive duality is linked neuroscientifically to the dynamic coupling of the brain's Executive Control Network (ECN) and Default Mode Network (DMN) \citep{beaty2016creative, andrews201813, chen2025dynamic, LUCHINI2023195}. While standard LLMs excel at tasks analogous to convergent thinking, they lack an intrinsic mechanism for the spontaneous, associative thought characterisatic of divergence. To address this deficit, we first sought to isolate a divergent generative process by fine-tuning on James Joyce's `Finnegans Wake'---a text that is itself a product of extreme associative thought. We then tested the capability of this process for conversational applications by integrating it into an instruction-tuned model, thereby creating and characterizing a family of specialized divergent tools.

\section{Methodology}
\subsection{Data and Training}
The foundational model of the family, TinyTim V1, was created by fine-tuning the `TinyLlama-1.1B-Chat-v1.0' model on the complete text of `Finnegans Wake' (approx. 1.5MB). The text was preprocessed into 100-word segments to preserve its associative structure, and the model was trained using a standard causal language modeling objective. TinyTim V1 has been available on HuggingFace for over a year \footnote{\href{https://huggingface.co/npc-worldwide/TinyTimV1}{hf.co/npc-worldwide/TinyTimV1}}, and has been downloaded more than 750 times in that time span.

Building on this foundation, we then developed TinyTim-V2-IT \footnote{\href{hf.co/npc-worldwide/tiny-tim-v2-1b-it}{https://huggingface.co/npc-worldwide/tinytim-v2-1b-it}} to test if the divergent style could be harnessed within a conversational framework. This instruction-tuned variant, based on `google/gemma-3-1b-it', was trained using TruthfulQA questions with answers rewritten in a Joycean style (100 epochs, learning rate 2e-4, LoRA rank 128), forcing the model to apply its divergent style to a convergent task.

In this work, we showcase and characterize both of these models.

\subsection{Evaluation Framework}
To quantify TinyTim's generative profile, we compared it against a pair of baseline models trained for coherent responses: `qwen3:0.6b' and `llama3.2'. We generated responses from each model using a set of 10 creative prompts. Each response was evaluated using syntactic and semantic metrics:
\begin{itemize}
    \item Syntactic Metrics: Unique Word Ratio, Average Word Length, Token Diversity (Shannon entropy), and Sentence Complexity.
    \item Semantic \& Content Metrics: Semantic Similarity to the prompt (via sentence embeddings), Readability (Flesch-Kincaid Grade Level), and Sentiment (VADER compound score).
\end{itemize}

\section{Results}
The generative profiles of the TinyTim models and their baselines reveal a statistically significant and functionally distinct separation between divergent and convergent systems. From an initial set of generated samples, we analyzed a filtered dataset of 371 high-quality responses: TinyTim-V1 (n=73), TinyTim-V2-IT (n=100), `llama3.2' (n=98), and `qwen3:0.6b' (n=100). A Kruskal-Wallis test confirmed a significant difference ($p < .0001$) among the model groups for all metrics, with subsequent pairwise tests showing that both TinyTim models were statistically distinct from the baselines on every metric.

\begin{figure}[h!]
  \centering
  \includegraphics[width=\columnwidth]{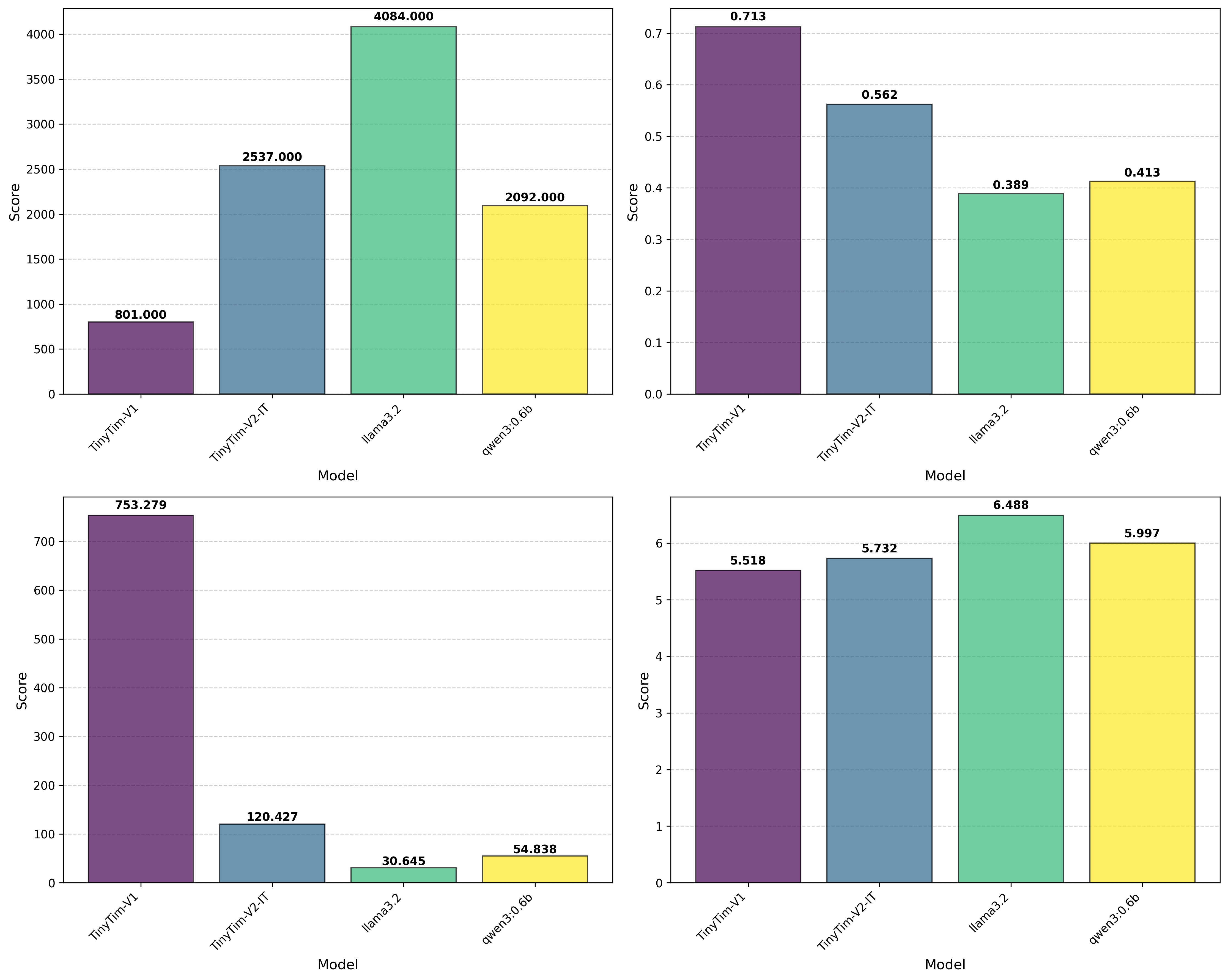}
  \caption{Measures of lexical sophistication. The TinyTim family, particularly V1, are clear outliers in lexical invention (Hapax Ratio, Yules K), distinguishing their creative generation from the large-vocabulary retrieval of baselines.}
  \label{fig:lexical_bars}
\end{figure}

\begin{figure}[h!]
    \centering
    \includegraphics[width=\columnwidth]{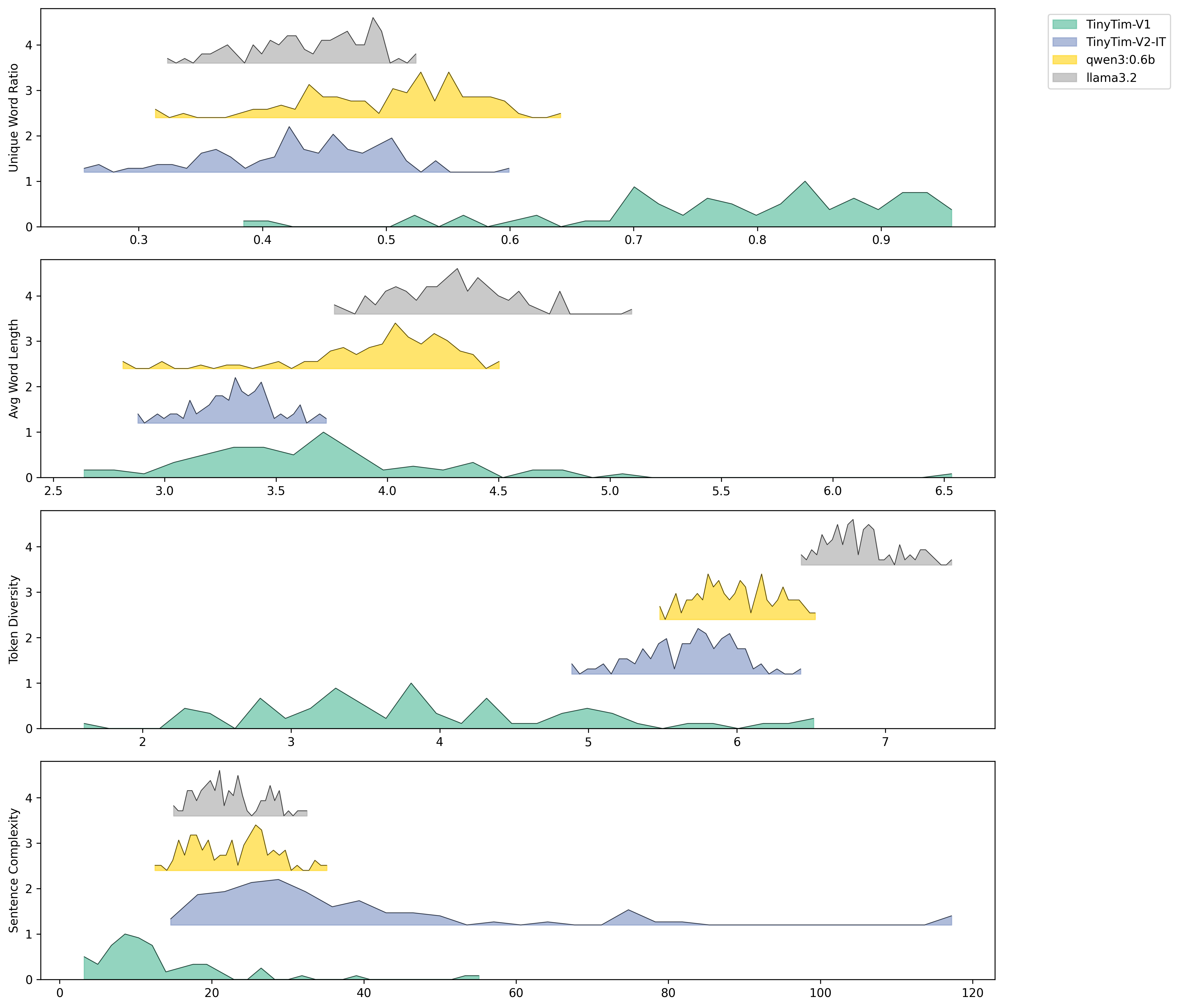}
    \caption{Ridge plots showing the distribution of scores for four primary metrics. The baseline models show tight, predictable distributions. Both TinyTim models exhibit extreme variance and long tails, indicative of a divergent generative process. The x-axis represents the score for the corresponding metric.}
    \label{fig:ridge_plots}
\end{figure}

\begin{figure}[h!]
  \centering
  \includegraphics[width=\columnwidth]{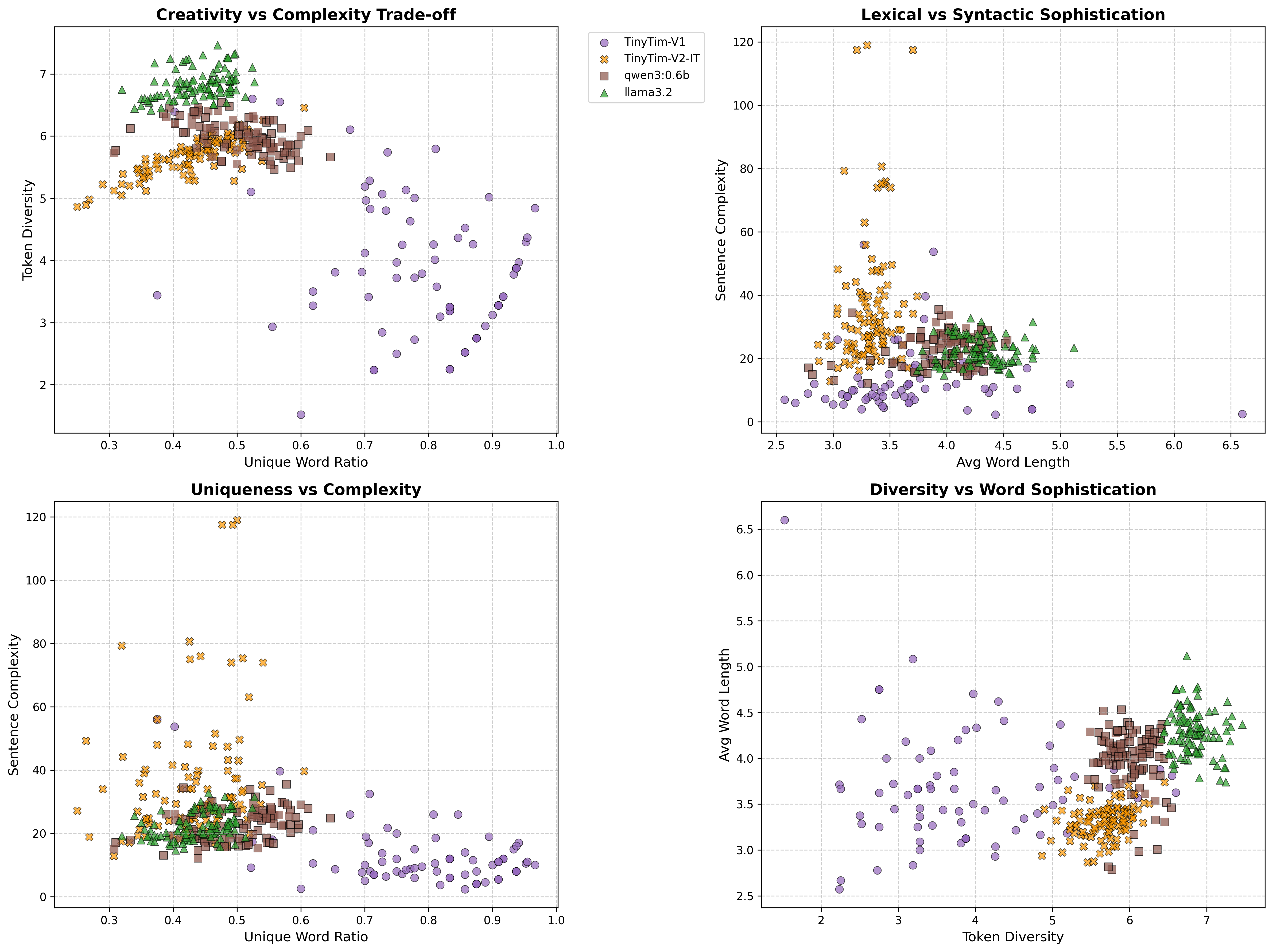}
  \caption{Scatter plots showing relationships between primary metrics. The top-left panel highlights the fundamental trade-off between Token Diversity and Unique Word Ratio, clearly separating the divergent strategy of the TinyTim family from the convergent strategy of the baseline models.}
  \label{fig:scatter_plots}
\end{figure}

Analysis of lexical sophistication reveals the core of the divergent modification. As shown in Figure \ref{fig:lexical_bars}, while a model like `llama3.2` possesses a vast vocabulary (4084 unique words in our sample), the TinyTim family demonstrates radical lexical invention. The foundational model, TinyTim-V1, achieves a Yule's K score—a robust measure of lexical richness—of 753.3, a value over twenty-four times greater than `llama3.2` (30.6). The instruction-tuned TinyTim-V2-IT, while more constrained, still registers a Yule's K of 120.4, quadruple that of the baselines. This shows that the TinyTim models are not simply retrieving from a large vocabulary; they are lexical inventors.
The distributional profiles of the models, shown in Figure \ref{fig:ridge_plots}, highlight the difference between convergent and divergent systems. The baseline models (`llama3.2', `qwen3:0.6b') produce outputs with tight, narrow, and predictable distributions for all metrics. This is the statistical signature of a reliable, convergent system optimized for consistency.
In contrast, the distributions for both TinyTim-V1 and V2-IT are extremely wide and heavily skewed. This is most apparent in Unique Word Ratio, where V1's distribution is nearly double that of any baseline, and in Sentence Complexity, where V2-IT produces outputs ranging from short fragments to sentences over 200 words long. This high variance serves as an indicator of their function as `unpredictable' generators.
The relationships between metrics further illuminate the different generative strategies. Figure \ref{fig:scatter_plots} shows the correlations between key metrics. The most significant finding is in the top-left panel, which plots Token Diversity against Unique Word Ratio. The baseline models form a tight cluster in the high-diversity, low-uniqueness space. They draw from a vast, general vocabulary (high Token Diversity) but exhibit more repetition within any single response (lower Unique Word Ratio). TinyTim-V1 occupies the opposite corner, forming a dispersed cloud in the high-uniqueness, low-diversity space, confirmed by the strong negative correlation for the full dataset ($r = -0.707$). TinyTim-V2-IT aligns more with the baseline models in this view, but diverges from them in sentence complexity and average word length. These results illustrate the core paradox of their capabilities: each individual response is highly novel (high UWR), but the overall pool of words they use is highly specialized and constrained to their training data (low Token Diversity). This trade-off is the defining feature of their specialized, divergent function.

\subsection{Benchmark Performance as a Confirmation of Divergence}

The stability of this divergent cognitive style was confirmed by testing TinyTim-V2-IT on a factual benchmark. It achieved only 52\% accuracy on AI2 ARC-Easy, compared to baselines: `llama3.2' (91\%), `qwen3:0.6b' (89\%), and `gemma3:1b' (80\%). The model's characteristic failure patterns included incoherent language, rambling responses, and hallucinated content, all evidence of its Joycean training overriding the instruction to be factual. This catastrophic failure is not a bug, but the critical piece of evidence proving that the divergent mode is a deep-seated, stable property of the model that resists convergence. It demonstrates a functional trade-off: the model sacrifices factual reliability to preserve its core generative style.

\section{Discussion}
The quantitative results show that fine-tuning on experimental literature produces a model with a statistically distinct, divergent generative profile. This moves the assessment of AI creativity beyond subjective claims and toward a measurable, multi-faceted profile.
\subsection{Challenging Parsimony in Creative Systems}
The principle of parsimony often guides the development of scientific models. However, recent work suggests that overly simplistic models can be insufficient for complex phenomena, and that complexity can be essential for scientific progress \citep{dubova2025}. Standard LLMs are inherently parsimonious; they seek the most probable, and thus often simplest, path through semantic space. `Finnegans Wake' is an anti-parsimonious text. By training on it, TinyTim learns a complex, non-minimal generative function. This demonstrates that for tasks requiring creative exploration, a non-parsimonious model trained on complex data may be more valuable than a simple one, as it is better equipped to navigate the high-complexity spaces where novel ideas reside \citep{boden1998creativity, holland1992genetic}. This aligns with findings that unguided, random exploration can be superior to rigid, theory-driven investigation \citep{dubova2022against, musslick2025automating}. Furthermore, when paired with a convergent model, outputs from TinyTim can be harnessed as part of an integrated system of exploration, positioning it as a powerful tool for automated discovery.
\subsection{Implications for Human-AI Collaboration}
The utility of TinyTim lies in its ability to generate novel linguistic fragments and non-obvious connections, fostering the conditions for serendipity \citep{mccay2015investigating}. This raw material requires interpretation. The user's role shifts from querying for an end to engaging in a co-creative, human-in-the-loop process \citep{hmcf2025, labelyourdata2025}. This process is analogous to Conceptual Blending, where new meaning emerges from the integration of disparate mental spaces \citep{fauconnier2003conceptual}. The model provides the disparate spaces; the human---or an alternative convergent LLM\footnote{The way Large Language Models interpret meaning in words appears to reproduce the contextuality effects observed in humans, see Agostino et al. (2025) on quantum semantics \cite{agostino2025quantum} and Wu et al. (2024) on the relative and sequential placement of words matters to LLMs and \cite{bruza2023} to read a recent description of the process observed in cognitive experiments.}---provides the interpretive act. A system that allows a user to invoke a `divergent agent' makes this creative strategy an accessible part of the collaborative process\footnote{Indeed, TinyTim has already played a role in powering creative work: \href{https://www.amazon.com/dp/B0DMWPGV18}{\textit{Don't turn on the sun} by giacomo catanzaro
}}, enhancing the system's ability to respond to context in a useful manner \citep{lampinen-etal-2022-language}. 
\section{Conclusion}
This work has demonstrated that the generative properties of a language model can be fundamentally reshaped through targeted fine-tuning on radically unconventional literature. By analyzing the TinyTim family of models, we have established a set of core findings regarding the engineering of specialized tools for augmenting current AI systems with more creative capabilities. Our major conclusions are the following:
\begin{enumerate}
    \item Fine-tuning on experimental literature quantitatively demonstrates that a language model's intrinsic generative bias can be shifted from a convergent to a divergent cognitive style, as evidenced by a statistically significant increase in lexical invention and output variance.
    \item The generative profile of the TinyTim family demonstrates their adherence to the uniqueness of their source material. Even in the instruction-tuned version, the characteristic style and inventiveness persist while functionally performing the tasks of responding to user prompts. Though its performance on the benchmark is degraded compared to its baseline `gemma3:1b' model (76\% versus 50\%), it still performs better than the `gemma3:270m' model (26\%) and thus motivates further fine-tuning and training on larger local models like `gemma3:27b' as well as the notion of a TinyTim-based reasoning model.
    \item The models' function as a generator of high-variance, low-coherence semantic material implies a new paradigm for human-AI interaction, shifting from a query-response dynamic to a co-creative partnership where the AI tool forces `out-of-the-box' thinking. Specifically, the conversational capabilities of TinyTim-V2-IT demonstrate that this co-creative partnership can be engineered into AI systems that assist not by providing answers, but by forcing a deeper, more divergent engagement with the problem itself.
\end{enumerate}

\bibliographystyle{plain}
\bibliography{references}
\end{document}